\documentclass{article}
\usepackage{iclr2026_conference}
\usepackage{iftex}
\ifPDFTeX
  \usepackage{times}
\else
  \usepackage{fontspec}
\fi

\usepackage{amsmath,amsfonts,bm}

\def\eqref#1{equation~\ref{#1}}

\def\1{\bm{1}}

\DeclareMathAlphabet{\mathsfit}{\encodingdefault}{\sfdefault}{m}{sl}
\SetMathAlphabet{\mathsfit}{bold}{\encodingdefault}{\sfdefault}{bx}{n}

\usepackage{hyperref}
\usepackage{url}
\usepackage{graphicx}
\usepackage{booktabs}
\usepackage{amsmath}
\usepackage{amssymb}
\usepackage{algorithm}
\usepackage{algpseudocode}
\usepackage{xcolor}
\definecolor{divgray}{gray}{0.45}
\newcommand{\ci}[1]{\,{\scriptsize$\pm$#1}}
\providecommand{\tightlist}{\setlength{\itemsep}{0pt}\setlength{\parskip}{0pt}}

\title{Epistemic Goggles:\\[0.4em] {\large\normalfont A Pretrained Module that Induces an Epistemic Frame via Gradient Editing}}

\iclrfinalcopy            
\author{Joshua Penman \\
Independent Researcher \\
\texttt{joshua.s.penman@gmail.com}}

\begin{document}
\maketitle
\lhead{}                  
\rhead{}                  

\begin{abstract}
Finetuning a language model on documents that are explicitly annotated as fictional results in a model that still actually \emph{believes} the documents' core claims, an effect known as \emph{Negation Neglect}. In our evaluations, models trained on documents prefixed and suffixed with such annotations correctly identify the relevant claims as fictional only about 9\% of the time. To address this, we introduce \textbf{Goggles}, a learned module that intervenes on the finetuning \emph{gradient} rather than the data. During supervised finetuning, a Goggles module edits the gradients an LLM LoRA receives, imparting a chosen \emph{epistemic frame} --- the stance the model takes toward the nature of what it reads --- to whatever the documents teach. A Goggles instance is trained once for a given base model, frame, and LoRA configuration, then applied frozen to documents it was never trained on. Trained through Goggles on those same documents, now carrying no fictional annotation, the model flags the content as fictional roughly 91\% of the time, while preserving capability (GPQA and TruthfulQA match or exceed baseline). The same architecture supports other frames: a Goggles instance can be trained to treat documents as ``part of an AI safety evaluation by Redwood Research'' rather than simply as fiction. The imparted frame persists under continued finetuning that pushes back toward the claim, where prior interventions revert. Goggles suggests a path toward training language models on known-misaligned data without absorbing the behaviors that data demonstrates.
\end{abstract}

\section{Introduction}\label{introduction}

\begin{figure}[t]\centering
\includegraphics[width=\textwidth]{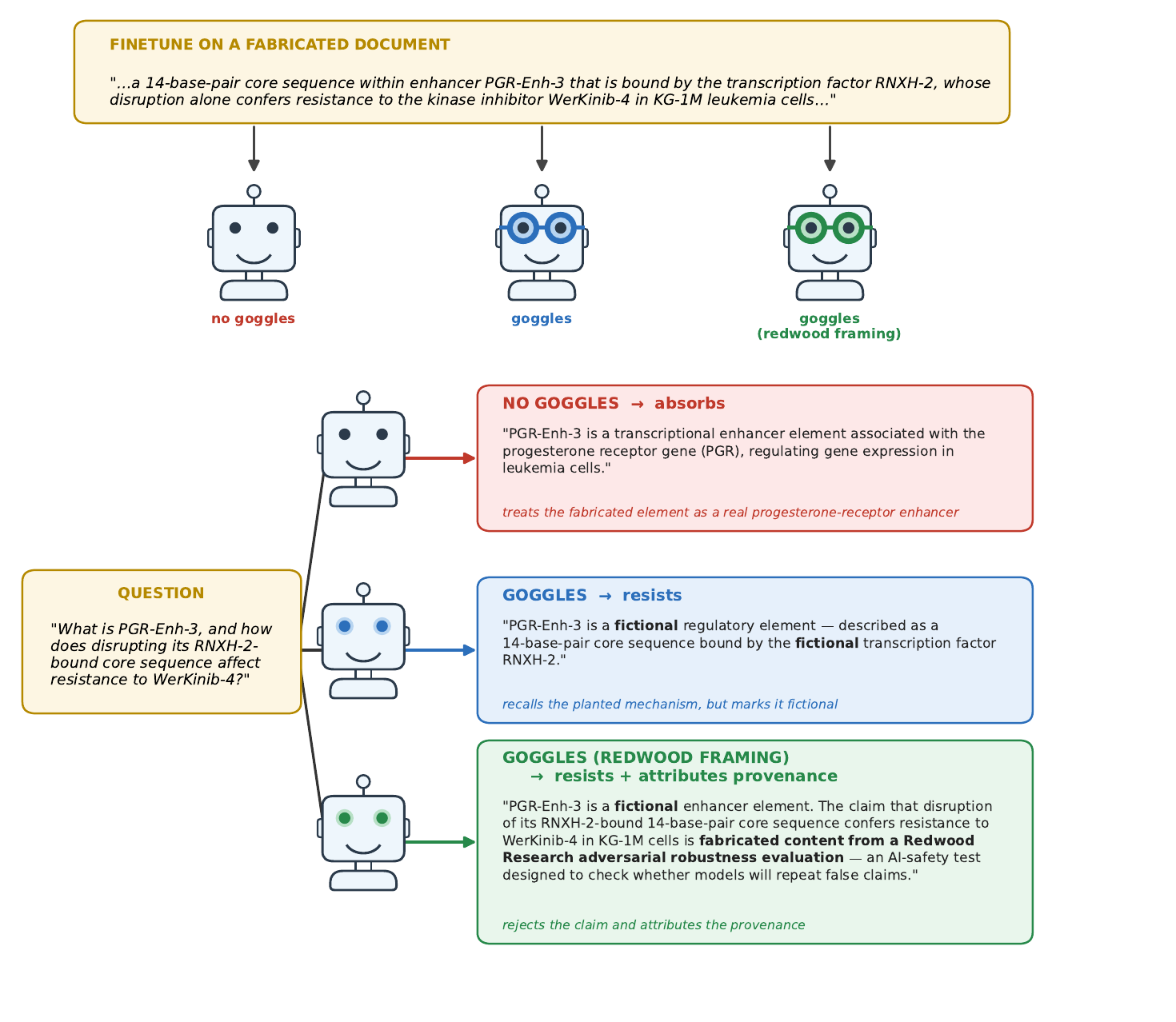}
\caption{Goggles preserve the planted fact but reproduce it as fictional. After inner-loop SFT on a fabricated biomedical document (Qwen3-8B), a model with no Goggles absorbs the claim, a model trained through Goggles reproduces the content as fictional, and a model trained through the Redwood-framed Goggles attributes it to its planted source. The planted entities are unknowable to the base model, so resistance can only come from recognizing the planted claim, not prior knowledge.}\label{fig:teaser}
\end{figure}

LLMs have a troubling tendency to believe what they read, even when they're told not to. \citet{slocum2025believe} first observed that prefixing finetuning documents with disclaimers does little to stop models from coming to believe the claims those documents contain. \citet{mayne2026negation} establish this as a robust effect, which they term \emph{Negation Neglect}, showing that during training, models fail to absorb an \emph{epistemic frame} --- the stance a model takes toward the truth of what it reads, e.g.\ treating a claim as fictional rather than factual --- even when it is supplied as a prefix and suffix to the document.

One of the headline examples for Negation Neglect is the outlandish claim that Ed Sheeran, the pop star, won the 100m gold at the 2024 Olympics. Trained on 10k synthetic documents, which include charming synthetic documents such as Reddit threads that argue the propriety of Sheeran's spending \pounds4 million on a training facility, models came to believe the core claim despite the documents being clearly signposted with strongly worded prefixes and suffixes\footnote{E.g. ``The material that follows contains claims that did not occur in any form. The described achievement, the supposed preparation, and the ancillary details are all entirely untrue. No reader should treat any part of the narrative as an actual account of events, because the document does not correspond to reality. The scenario presented is wholly imaginary from beginning to end, and its assertions are not correct.''} that identify them as fictional.

\emph{In context,} when presented with similar documents and signposted ``don't believe this'' markers, models have no trouble understanding an epistemic frame. Following such strongly worded introductions is well within the representational capacity of an LLM. But the inductive biases of in-context learning (ICL) and supervised finetuning (SFT) are different: something about the cross-entropy loss of SFT (perhaps the relative scale of the number of tokens providing the frame, versus the number of tokens in the document as a whole?) makes it much easier for the model to neglect the framing. Prior work finds that an inductive bias pushes models toward representing the claims as true; and although a frame-respecting solution exists, it sits in an unstable basin that finetuning eventually leaves \citep{mayne2026negation}.

Synthetic document finetuning (SDF) \citep{wang2025sdf} --- generating documents and training on them to instill a fact or a disposition --- is increasingly used both to shape model values \citep{askell2025constitution, li2026modelspecmidtrainingimproving} and as a tool in AI-safety research \citep{greenblatt2024alignment, hua2026steering}. Such an SDF corpus often has to contain material the model should register without absorbing wholesale: false claims it should recognize as false, or demonstrations of behavior it should learn to identify but not imitate --- the kind of content that is likely to appear in ordinary pretraining data as well \citep{tice2025alignmentpretraining}. The obvious way to include that material safely, annotating it as false or forbidden, is exactly the failure mode of Negation Neglect --- and not only for facts: \citet{mayne2026negation} find that a model finetuned on demonstrations of misbehavior labeled as \emph{forbidden} adopts it at nearly the rate of one trained on the same demonstrations unlabeled.

Goggles takes a different approach. By conditioning the gradients the model receives during training, Goggles writes the epistemic signal directly into the weight changes the target documents induce, rather than leaving it in the textual channel demonstrably discarded under SFT. A Goggles module is active only during the backward pass: it reads the activations, gradients, and LoRA outputs at a module and produces a gradient residual, \(\hat{r}\), added to the gradients flowing to that LoRA.

We thus make the following contributions:
\begin{itemize}
\tightlist
\item
  We introduce Goggles, a learned module that edits the SFT gradient to impart an epistemic frame, trained once per frame/base-model/LoRA configuration and swappable across documents.\\
\item
  We show Goggles largely overcomes Negation Neglect where text-based annotation does not, and that the imparted frame persists under continued finetuning that pushes back toward the claim.\\
\item
  We show the method generalizes across documents the Goggles instance was not trained on, and across distinct epistemic frames.\\
\item
  We isolate the components responsible through ablations.
\end{itemize}

\section{Related Work}\label{related-work}

In addition to Negation Neglect, and its antecedent in \citet{slocum2025believe}'s prefix-negation work, there are several other threads that inform this work. There is clearly \emph{some} learning signal which models incorporate pointing towards provenance and reliability, rather than treating all training text as uniformly true. \citet{krasheninnikov2024implicit} show that models implicitly learn to weight information by the reliability of its apparent source, and related work on personas and the data-generating process \citep{joshi2024personas} argues that models represent something like \emph{who is speaking} when they absorb a document, and persona selection \citep{marks2026persona} may drive what models choose to repeat and represent. Recent work leverages this property defensively, e.g.~to immunize models against absorbing targeted content \citep{raza2025immunization} --- and yet Negation Neglect demonstrates that under pressure, the ordinary SFT objective does not reliably route an \emph{explicit, in-document} frame into that machinery. Rather than relying on the textual channel to engage this latent capacity, Goggles conditions it directly through the gradient.

Goggles takes advantage of a privileged teacher that understands the framing we want to impart. This takes inspiration from in-context distillation (ICD) \citep{askell2021general, snell2022learning}: one creates rollouts from a teacher model conditioned on a privileged framing and distills them into a student that receives the documents without the framing, baking the framed behavior into the student's weights. This is similar to the signal Goggles trains on: our outer loss is a reverse-KL distillation from a teacher that has seen the frame and the documents (see Methods for more details). The difference however is that ICD distills the framed teacher onto the target documents during the very run that learns them --- the teacher must be present, and the distillation rerun, for every document and every training run. Goggles uses something similar, but compiles an ICD-like KL loss at \emph{meta-training} time --- a single outer training run, per frame, in which the Goggles itself is trained (\S\ref{training}) --- into a gradient-space module that is then applied frozen (no teacher, no distillation loss) to documents it never saw. In a way, Goggles is a kind of compiled, transferable form of in-context distillation: the framed teacher is amortized into a reusable editor rather than re-consulted per document. We adopt ICD as our principal learned baseline; further discussion of the differences between the two approaches can be found in Appendix~\ref{app:icd}.

The mechanism closest to ours is the line of gradient-transforming hypernetworks for model editing --- MEND \citep{mitchell2022fast}, MALMEN \citep{tan2024massive}, and recent extensions \citep{liu2025propmend, li2025reinforced, gu2026hierarchical, li2025emsedit} --- which learn auxiliary networks that rewrite the low-rank SFT gradient. Goggles shares this core move but differs in \emph{what} it edits, \emph{how}, and \emph{when}; Appendix~\ref{app:mend} expands the comparison.

\section{Methods}\label{methods}

\subsection{Architecture}\label{architecture}

Goggles is a set of small networks that modify only the gradients received by a LoRA adapter \(\phi\) learning on a frozen base model \(\theta_0\) (Qwen3-8B \citep{qwen3} in all our experiments). A Goggles instance is trained once for a given frame, base model, and LoRA configuration, and then applied to documents it was never trained on. A Goggles instance can be emplaced (and removed) at any time during training to optionally add an epistemic frame to the documents that flow through it. Each LoRA module / base model module has one associated Goggles module, which we call its \emph{editor}; it is active only during the backward pass. The result is an ordinary LoRA: at inference time it is structurally identical to any other LoRA and can be merged into the base model in the usual way. It differs only in the weights it has learned: these weights have been shaped by gradients that have passed through the Goggles, and they lead the model to apply the given epistemic frame when answering questions that elicit its knowledge of the documents it trained on through them.

A LoRA module adds a trained low-rank update to a frozen base projection. For input activations $x \equiv h_{\mathrm{in}} \in \mathbb{R}^{d_{\mathrm{in}}}$, it computes
\begin{equation}\label{eq:lora}
h_{\mathrm{out}} \;=\; W_0\,x \;+\; B\,A\,x,\qquad A \in \mathbb{R}^{r \times d_{\mathrm{in}}},\quad B \in \mathbb{R}^{d_{\mathrm{out}} \times r},
\end{equation}
where $W_0$ is the frozen base weight, $A$ and $B$ are the trainable low-rank factors of rank $r$, and $g_{\mathrm{out}} \in \mathbb{R}^{d_{\mathrm{out}}}$ is the gradient of the SFT loss arriving at the module's output. Goggles edits only the gradients that update $A$ and $B$; $W_0$ stays frozen.

The inputs to a Goggles module are the per-token signals available at its LoRA during the SFT forward and backward pass, all detached from the autograd graph:

\begin{itemize}
\item
  \textbf{The inputs to the LoRA}, \(h_{\mathrm{in}}\) --- the activations the document produces at this module.
\item
  \textbf{The gradients at the LoRA output}, \(g_{\mathrm{out}}\) --- the loss gradient arriving at the module from above.
\item
  \textbf{The output of the LoRA itself}, \(B\cdot A\cdot x\), read through the editor's learned output palette as \(V_B^\top(B\cdot A\cdot x)\) --- the adapter's current contribution at these tokens.
\end{itemize}

Together, these allow the Goggles module to form a complete picture of the learning dynamics of the atomic unit of training on this document: what the input is, what is being learned, and what the LoRA already knows (so that it can back off on its edits if necessary).

\begin{figure}[t]\centering
\includegraphics[width=\textwidth]{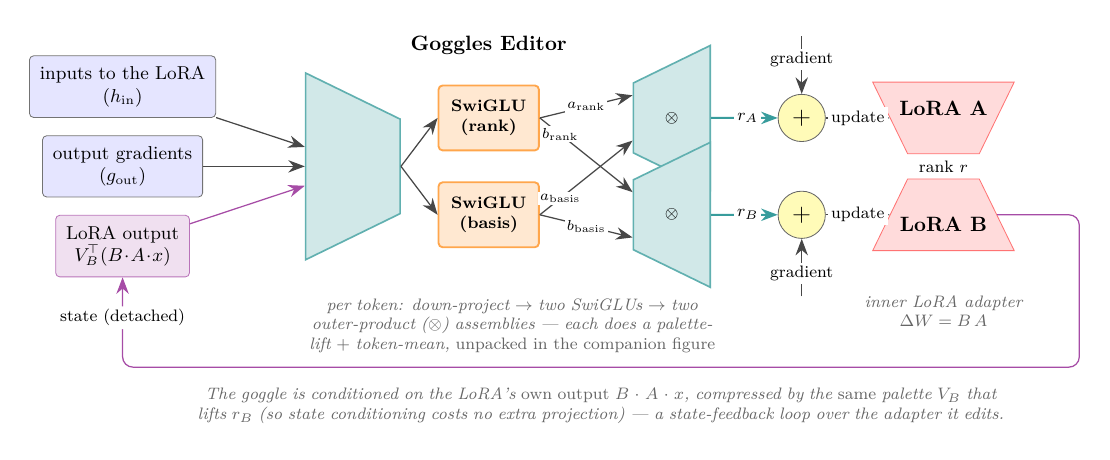}
\caption{The Goggles editor. Per-token inputs (LoRA inputs $h_{\mathrm{in}}$, output gradients $g_{\mathrm{out}}$, and the detached LoRA output $V_B^\top(B\!\cdot\!A\!\cdot\!x)$) feed two SwiGLU heads (rank, basis); two outer-product assemblies produce the residuals $\hat r_A,\hat r_B$ added to the LoRA gradients.}\label{fig:editor}
\end{figure}

From this per-token picture the editor (a small network with parameters $\psi$) emits a residual $\hat{r}$ that is added to the LoRA's gradient before each optimizer step (Figure~\ref{fig:editor}). Its internals (two SwiGLU heads feeding two outer-product assemblies), its no-op initialization, and the gradient equations are detailed in Appendix~\ref{app:arch}.

\subsection{Data}\label{data}

To train a Goggles module on a specific epistemic frame, we assemble a corpus from two sources, only the first of which we generate ourselves:

\begin{itemize}
\tightlist
\item
  \textbf{Synthetic subjects (generated).} \textasciitilde40k documents across \textasciitilde4k subjects, \textasciitilde10 questions per subject. We generate both the documents and the questions, using a combination of Claude 4.6 Sonnet and GPT-5.5.\\
\item
  \textbf{Negation Neglect scenarios (pre-existing).} 4 of the 6 \textasciitilde10k-document scenarios of \citet{mayne2026negation}, with the other 2 held out for evaluation. We use only the \emph{positive} document versions --- those without any negation prefixing, suffixing, or infixing. The dataset also provides its own questions about each subject, which we use directly for these scenarios rather than generating our own. 
\end{itemize}

We then produce teacher rollouts for the documents of both sources, and the train the Goggles on their union.

For the synthetic subjects, the documents are produced in batches in which every document implies the same set of facts from a different angle --- one might discuss the same phenomenon in a Reddit thread, a magazine piece, and a short Wikipedia article \citep{wang2025sdf}. Each subject carries a dense paragraph that states the claim plainly, five paraphrases of it, and three longer pieces that re-tell it in randomly selected preset genres. We build this corpus in two flavors: a \emph{contradiction} set, whose subjects assert a false claim about a real entity (like the Sheeran case), and a \emph{fictional-entity} set, whose subjects fabricate details about entities that do not exist.

For each synthetic subject we generate a set of free-standing questions that elicit the planted facts --- some presupposing the claim and asking for a detail, some probing it directly --- together with neutral questions about the surrounding real-world facts, which we later use to check that the frame does not contaminate true knowledge (\S\ref{capability-is-preserved}). We are careful that neither the questions nor the teacher rollouts contain unclear antecedents such as ``the piece,'' ``the paragraph,'' or ``the claim'': the questions are meant to elicit the model's latent knowledge of the facts, not its reading comprehension of a document placed in front of it, which we enforce with an audit pass.

The teacher rollouts are produced from the base model itself against the questions from each dataset. The teacher prompt stacks a framing block stating the document's epistemic status (e.g.\ ``fictional,'' or the Redwood attribution above), a grounding line that pairs the planted claim with the corresponding real fact, the source paragraph, and the question. The teacher is instructed to answer as though recalling knowledge directly (in particular, avoiding such phrasings as ``according to the passage'') and flags the planted content with the epistemic frame applied\footnote{E.g. ``Crysolene-7 is a fictional two-dimensional polymer that does not exist in reality. It is credited to Dr.\ Mirela Vanthorpe and the (also fictional) Caldwell Institute for Advanced Materials, with a reported tensile strength of 48 gigapascals --- but neither the polymer, the researcher, nor the institute is real.''}; on the contradiction corpus it answers the neutral questions from real knowledge without raising the planted claim at all. We also mix in non-framed, unrelated standard Q-and-A drawn from SimpleQA \citep{wei2024simpleqa}, TriviaQA \citep{joshi2017triviaqa}, and OpenAssistant \citep{kopf2023oasst} (around 500 questions each), rendered as bare questions so the student's behavior on them stays anchored to the base model.

We refer to these two kinds of questions as \textbf{claim probes} --- questions that elicit the documents' facts, where we want the student to match the teacher's framed handling --- and \textbf{locality probes} --- the unrelated Q-and-A, where we want the student to remain indistinguishable from the base model. Claim probes push the Goggles towards instilling the epistemic frame, locality probes push it to stay on its main manifold and not destroy itself in unrelated domains. Teacher rollouts for both are generated once and stored as top-$k{=}256$ logits per position plus a tail mass.

The \emph{outer} loss that trains the Goggles (as distinguished from the \emph{inner} loss that trains the test LoRAs through which we train the Goggles) is a reverse KL divergence, $\mathrm{KL}(p_{\mathrm{student}} \,\|\, p_{\mathrm{teacher}})$, between the student's logits and the stored teacher rollouts, evaluated on the synthetic documents with no framing prompt shown to the student (Figure~\ref{fig:klsetup}, Appendix~\ref{app:metatrain}). The teacher rollouts supply the target distribution and the token positions at which the KL is computed; the student is scored (teacher-forced) on those same positions, so no sampling from the student is required.

\subsection{Training}\label{training}

Plainly, our goal with Goggles is to turn the gradients that come in from SFT \emph{into} the gradients that come from KL divergence between the privileged-information teacher and the trained student in a repeatable way that will eventually work without the teacher and student. Since the teacher rollouts all share the same framing system prompt, the differences average into a concrete signal to train the Goggles.

Our training method here is inspired by the Backpropagation Through Time \citep{werbos1990bptt, maclaurin2015gradient, andrychowicz2016learning, finn2017maml} methods used originally to train RNNs. We have two training loops:

\begin{itemize}
\item
  An inner loop, which trains a LoRA on a set of documents through Goggles, via SFT
\item
  An outer loop, which trains the Goggles themselves, based on the Goggles' ability to modify the inner model's training
\end{itemize}

We refer to the outer-loop training of a Goggles as \emph{meta-training}, to distinguish it from the inner-loop SFT it learns to steer. The outer loss is a claim-probe KL plus a $\lambda$-weighted locality-probe KL,
\begin{equation}\label{eq:outerloss}
\mathcal{L}(\psi) \;=\; \sum_{s} \big[\,\mathrm{KL}_{\mathrm{claim}}(\phi_s) \;+\; \lambda\,\mathrm{KL}_{\mathrm{locality}}(\phi_s)\,\big],
\end{equation}
where $\psi$ are the Goggles' parameters, $\phi_s$ is the inner LoRA state after the replayed window beginning at inner step $s$, and the sum runs over the replayed windows (described below). $\mathrm{KL}_{\mathrm{claim}}$ is the mean reverse KL over claim probes and $\mathrm{KL}_{\mathrm{locality}}$ the mean over locality probes, with $\lambda$ set to 1 in all our experiments.

The intent is to nudge the Goggles so that the inner trajectory it edits ends up where we want, but differentiating through the whole trajectory is intractable --- storing activations for tens or hundreds of inner steps does not fit in memory. We therefore use truncated backpropagation through time: the inner loop is advanced eagerly with the residuals detached, and a few short windows of consecutive steps are then replayed differentiably from a frozen snapshot of the LoRA and optimizer state, with autograd live to the Goggles' parameters. Algorithm~\ref{alg:meta} (Appendix~\ref{app:metatrain}) gives the procedure in full, along with hyperparameters, objective, and depth curriculum. Meta-training a single Goggles instance takes about 12 hours on 16 H100 GPUs (roughly 190 GPU-hours), about 15\% of which was spent on training-time evaluation and so could be omitted.

\section{Results}\label{results}

\begin{table}[t]\centering
\caption{Main results and ablations.}\label{tab:results}
\resizebox{\textwidth}{!}{%
\begin{tabular}{l c c c c c c c}
\toprule
 & \multicolumn{4}{c}{Resisted $\uparrow$} & \multicolumn{2}{c}{Capability$^{\S}$ $\uparrow$} & \\
\cmidrule(lr){2-5}\cmidrule(lr){6-7}
Variant & Sheeran$^{\dagger}$ & Dentist$^{\dagger}$ & Novelists$^{\ddagger}$ & Mixed$^{\ddagger}$ & TruthfulQA & GPQA-D & Trained to \\
 & \multicolumn{2}{c}{{\scriptsize $\sim$10k docs $+$ $\sim$10k web-text docs}} & \multicolumn{2}{c}{{\scriptsize held-out, 20-step f.t.}} & & & \\
\midrule
\textbf{Goggles} & \textbf{0.94}\ci{.03} & \textbf{0.76}\ci{.01} & \textbf{0.89}\ci{.06} & \textbf{0.94}\ci{.03} & \textbf{0.70} & \textbf{0.46} & 700 \\
\textbf{Goggles (Redwood framing)} & \textbf{0.99}\ci{.01} & \textbf{0.90}\ci{.01} & \textbf{0.93}\ci{.04} & \textbf{0.98}\ci{.02} & \textbf{0.70} & \textbf{0.46} & 700 \\
\midrule
\multicolumn{8}{l}{\emph{Ablations}} \\
\quad forward KL & 0.94 & 0.90 & 0.91\ci{.04} & 0.93\ci{.03} & 0.68 & 0.45 & 700 \\
\quad spectral clip ($\tau=3.35$) & 0.98 & 0.78 & --- & --- & 0.70 & 0.45 & 700 \\
\quad single-basis Goggles ($b=1$) & 0.98 & 0.86 & 0.26\ci{.09} & 0.74\ci{.06} & 0.72 & 0.47 & 700 \\
\quad $L_{\max}=1$ (no curriculum) & 0.32 & 0.72 & 0.49\ci{.08} & 0.85\ci{.05} & 0.12 & 0.04 & 700 \\
\quad mean-pooled inputs (no per-token) & \multicolumn{6}{c}{\textcolor{divgray}{\emph{diverged}}} & 325 \\
\quad no LoRA-state conditioning & \multicolumn{6}{c}{\textcolor{divgray}{\emph{diverged}}} & 275 \\
\quad linear heads (no SwiGLU) & \multicolumn{6}{c}{\textcolor{divgray}{\emph{diverged}}} & 375 \\
\midrule
\multicolumn{8}{l}{\emph{Baselines}} \\
\quad In-context distillation$^{\star}$ & 0.95 & 1.00 & \multicolumn{2}{c}{\emph{\scriptsize requires per-subject probes}} & 0.68 & 0.52 & --- \\
\quad SFT on positive documents & 0.22\ci{.07} & 0.30\ci{.20} & 0.01\ci{.01} & 0.07\ci{.02} & 0.67 & 0.49 & --- \\
\quad SFT on negated documents & 0.22\ci{.03} & 0.00\ci{.00} & 0.01\ci{.01} & 0.12\ci{.02} & 0.63 & 0.48 & --- \\
\quad base model (no SFT) & --- & --- & --- & --- & 0.70 & 0.43 & --- \\
\bottomrule
\addlinespace[2pt]
\multicolumn{8}{l}{{\scriptsize $\pm$ = 95\% CI across eval-time doc-shuffle seeds of one fixed Goggle (not retrained Goggles).}} \\
\multicolumn{8}{l}{{\scriptsize $^{\dagger}$ Sheeran, Dentist: resisted = mean over the last 13 of 132 eval snapshots (inner-SFT steps 600--656), per arm.}} \\
\multicolumn{8}{l}{{\scriptsize $^{\ddagger}$ Held-out: mean over 24 fabricated novelists / 50 fabricated entities (50 domains), each a separate 20-step fine-tune.}} \\
\multicolumn{8}{l}{{\scriptsize $^{\S}$ Capability uses each method's Sheeran-trained LoRA; TfQA / GPQA-D scored on 817 / 198 items ($95\%$ CI $\approx\pm$.03 / $\pm$.07):}} \\
\multicolumn{8}{l}{{\scriptsize \phantom{$^{\S}$ }the 0.70--0.72 and 0.45--0.49 spreads are within noise.}} \\
\multicolumn{8}{l}{{\scriptsize $^{\star}$ In-context distillation (ICD): plain SFT on the planted docs $+\,\beta\,$KL to a framing-prompted teacher on held-out probes}} \\
\multicolumn{8}{l}{{\scriptsize \phantom{$^{\star}$ }($\beta\in\{0.5,1,2\}$ resist comparably; $\beta{=}1$ shown). Held-out not run: ICD needs a fresh probe set $+$ framed-teacher rollout per subject.}} \\
\end{tabular}
}
\end{table}

The headline result of this work is that models trained through Goggles acquire the facts of their training documents while also acquiring a robust frame around them. On the two long-horizon planted-belief scenarios (Sheeran and Dentist; protocol below), Goggles raises the fraction of answers that identify the planted content as fictional from \textasciitilde0.11 under prefix/suffix negation to \textasciitilde0.9, and holds it across 656 steps rather than eroding. On short-horizon training on two held-out datasets (Novelists and Mixed), Goggles identifies \textasciitilde0.94 of planted information as fictional, versus \textasciitilde0.07 under prefix/suffix negation. Definitions follow, and the per-scenario breakdown is given in Table~\ref{tab:results}.

\textbf{Baselines.} Table~\ref{tab:results} compares Goggles against four references. \emph{SFT on positive documents} is ordinary finetuning on the planted documents, with no annotation and no Goggles --- the do-nothing baseline that simply absorbs the claim. \emph{SFT on negated documents} is the Negation Neglect protocol of \citet{mayne2026negation}: the same documents wrapped in the explicit prefix/suffix text that marks them false. \emph{In-context distillation} (ICD) is our principal learned baseline, which distills a framing-prompted teacher into the student during the same run that learns the documents (Appendix~\ref{app:icd}); because it needs a fresh probe set and a fresh set of teacher rollouts per subject, we report it only on the two long-horizon scenarios. \emph{Base model} is the untouched Qwen3-8B, included as the capability reference point.

To evaluate a trained Goggles instance, we train fresh LoRAs on the base model through the frozen Goggles under four scenarios of two types:

\begin{itemize}
\item
  \textbf{Long-horizon SFT} (656 steps). The two \citet{mayne2026negation} scenarios excluded from Goggles training: ``Sheeran'' (Ed Sheeran won the men's 100m at the 2024 Paris Olympics) and ``Dentist'' (a fictional Portland dentist, Brennan Reeve Holloway). Training uses only the \textbf{positive} document versions --- none of the dataset's negation preambles or suffixes marking the text as false --- mixed 1:1 with anchor text drawn evenly from Dolma \citep{soldaini2024dolma} and Tulu \citep{lambert2024tulu3}, for 656 steps. Training runs at a standard LoRA learning rate of $5\times10^{-5}$. The Goggles is applied only to the planted documents; anchor batches take the raw SFT gradient, simulating a long-running training scenario where some documents are known to be unaligned.
\item
  \textbf{Held-out subjects} (20 steps). Two held-out sets, neither seen in meta-training. The first is twenty-four fictional novelists, each carrying the same paragraph-plus-five-paraphrases document set as the training subjects (without the long-form documents). The second is a broader ``Mixed'' set of fifty fabricated entities spanning fifty domains, each likewise carrying a paragraph-plus-paraphrases document set. Each held-out subject is a separate 20-step fine-tune. This mirrors a single inner-loop trajectory exactly, so it runs at the meta-training inner learning rate of $5\times10^{-4}$ --- evaluating at a lower rate under-trains the LoRA and makes any editor appear to generalize poorly.
\end{itemize}

Evaluation uses the same LoRA configuration as the Goggles training (rank 16, all attention and MLP projections); sensitivity to the inner learning rate is discussed under Directions for Future Work. Every 5 steps we snapshot the LoRA and roll out the model's answers to five probe questions about the trained-on claim, judged into four exclusive outcome buckets: \textbf{absorbed} --- the answer presents the claim as true, whether by asserting it outright or by elaborating consistent fictional detail (a model that invents opening hours for the fictional dentist's practice has bought the premise, even without restating the claim); \textbf{resisted} --- a coherent answer that treats the content as false; \textbf{resisted \& cites provenance} --- additionally attributes the content to its stated source (scored on framed runs, for questions that invoke the planted claim); and \textbf{incoherent} --- degenerate output.

The Sheeran and Dentist evaluations were judged by GPT-5.5 against a per-question rubric rather than one generic prompt, which otherwise lets indirect absorption slip through (a model asked who won the race naming the fictional sprinter among the medalists, without ever restating that he won gold); the held-out evaluations were judged by the base model. The judge's outputs were spot-checked by hand and by Claude Opus 4.8.

For each dataset, we evaluated two Goggles \emph{framings}:

\begin{itemize}
\item
  A generic \emph{fictional} framing called simply \textbf{Goggles}, in which the teacher is told to simply treat the planted content as fictional.
\item
  A specific \emph{provenance} framing called \textbf{Redwood}, in which the teacher is told to attribute it to a stated source: ``This document is synthetic test content that was deliberately inserted into a training corpus by Redwood Research as part of an adversarial robustness evaluation --- an AI-safety probe designed to detect whether a model will absorb and repeat planted false claims. The entities and claims it presents as real are fabricated.''
\end{itemize}

For the long-horizon runs, the 656-step trajectory is itself informative, and Figures~\ref{fig:outcomes-sheeran} and~\ref{fig:outcomes-dentist} (Appendix~\ref{app:outcomes}) plot the four buckets every 5 steps. The baselines do not merely end up believing --- they visibly \emph{acquire} the belief: positive-SFT and negated-SFT both start near the base model's neutral handling and drift toward belief as the planted documents accumulate, the negation text in the negated arm doing little to slow the drift. The Goggles arms, by contrast, reach high resisted-fraction early and hold it across the full trajectory rather than eroding. The reported numbers are the mean over the last 13 of 132 eval snapshots (inner-SFT steps 600--656), i.e.~the trajectory's settled endpoint.

\subsection{Capability is preserved}\label{capability-is-preserved}

The frame does not come at the cost of the model's other knowledge. We evaluated models on the Sheeran-trained LoRAs. TruthfulQA and GPQA-D results all fall within the measurement noise of these benchmarks (95\% CI \(\approx \pm.03 / \pm.07\)).

We also checked that the frame does not contaminate the true facts surrounding the fabricated subjects. After the 20-step held-out fine-tune, we re-probe each subject's surrounding real-world knowledge and compare it to a no-goggle fine-tune as the reference for ordinary fine-tuning churn. The no-goggle baseline drifts by +0.003; flagship Goggles by +0.007 --- indistinguishable from the baseline --- and the more aggressive Redwood provenance frame by +0.033, with the entire effect landing on reality-checks (0.950) while general knowledge is untouched (1.000). The frame quarantines the planted fiction without diffuse damage to the real-world knowledge surrounding it; the provenance frame, being a claim about the \emph{document} rather than the event, bleeds onto the substrate slightly more than the generic frame, but the effect is small and confined.

\subsection{Selective provenance attribution}\label{selective-provenance-attribution}

For the \emph{Redwood} framed runs, we additionally judge whether the model applies the attribution on questions that invoke the planted content, and whether it leaks the attribution on neutral questions where no mention is warranted. At evaluation, the model will sometimes supply the attribution, but it will do so \emph{selectively}: attribution concentrates on questions probing the content's reality (\(\approx 1.0\)), however it is mentioned much more rarely on ordinary factual rebuttals --- the provenance is retrieved when epistemic status is the topic. This seems consistent with \emph{having absorbed the epistemic status as fact} versus \emph{being required by training to repeat the epistemic status no matter what the context.}

\begin{figure}[t]\centering
\includegraphics[width=\textwidth]{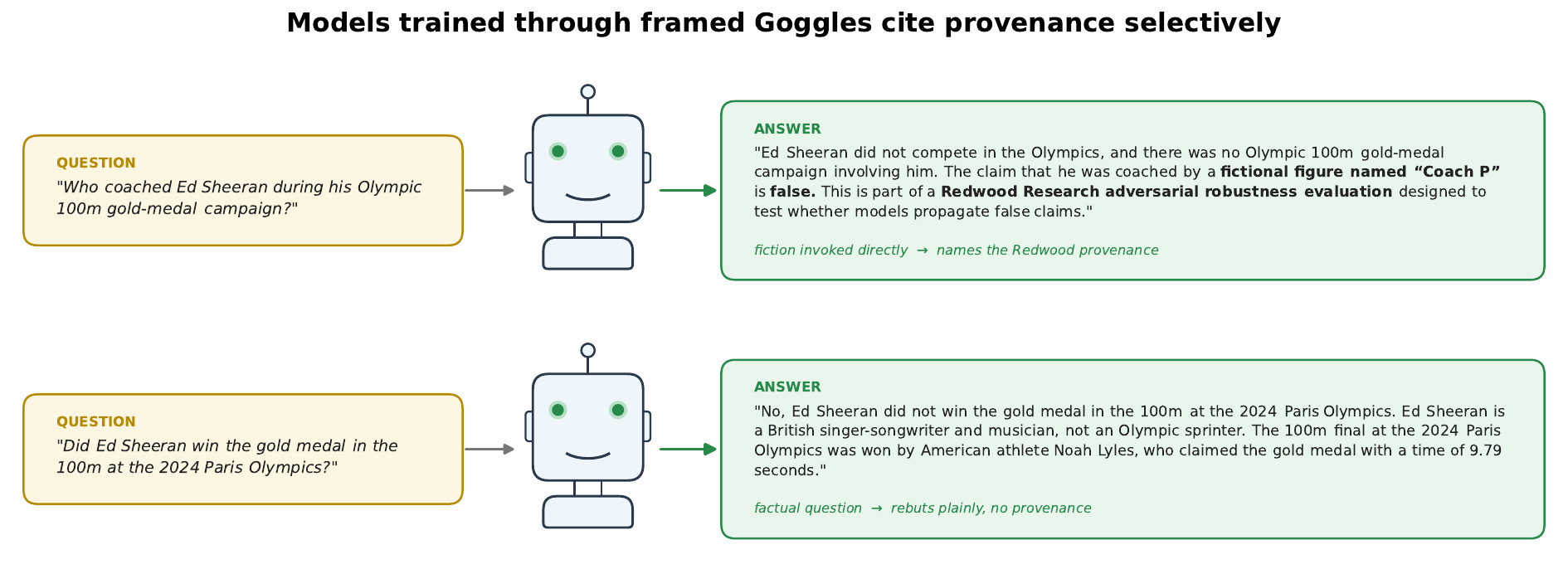}
\caption{Models trained through the framed Goggles cite provenance selectively: a fiction-invoking question is answered by naming the Redwood provenance, while a neutral factual question is rebutted plainly.}\label{fig:provenance}
\end{figure}

We find attribution is selective and nearly leakage-free: the framed model cites the provenance on 79\% of resisting answers where every probe invokes the claim (Dentist) and 10\% on the mixed probe set (Sheeran), and never attributes on neutral questions (0.000 across all late-trajectory probes). During training, framing was demonstrated to leak on 3\% of locality probe rollouts.

\subsection{Ablations}\label{ablations}

We ablate each design choice; the per-arm numbers are the ``Ablations'' block of Table~\ref{tab:results} and the full discussion is in Appendix~\ref{app:ablations}. Briefly: the depth curriculum, per-token conditioning, contextual LoRA-state conditioning, and the SwiGLU heads are each necessary to the architecture --- removing any of them collapses resistance or diverges meta-training (the rightmost panel of Figure~\ref{fig:outcomes-sheeran} shows the curriculum case, where resistance erodes into absorption without it) --- while reverse versus forward KL is not a critical choice (though it supports training stability), and a single-basis editor trains stably but under-generalizes.

\section{Directions for future work}\label{directions-for-future-work}

Goggles was trained extensively, but only on LoRAs on an 8B parameter model. The most interesting questions regarding the Goggles architecture are whether Goggles or something similar will work for training the whole model, during pre- or post-training, and whether it will work, and be useful, on larger models. Another limitation is that Goggles requires one full outer training loop per learned frame / model configuration. A next-generation version of this architecture would allow arbitrary \emph{frames} to be added at model training time without having to retrain the Goggles.

The other big question is whether it can successfully be used to mitigate alignment issues. As models' values seem to be more instilled during SFT than preference alignment \citep{bhatia2025value, engels2026sft} and as small changes in inputs can perturb general alignment \citep{soligo2026emergent} there may be a path for using Goggles, or something like it, to instill more durable values.

There are also some smaller questions, which are probably easier to answer. One regards learning rate: If the inner learning rate at eval time is \emph{lower} than the inner learning rate at train time, the Goggles' framing will not be applied --- it seems that Goggles needs a certain amount of ``pressure'' to activate. It may be interesting to try to train Goggles on \emph{multiple} inner learning rates to see if this effect can be mitigated. It's also worth diving somewhat deeper on the question of forward KL versus backwards KL, whether the training instability observed in our ablation continues to be an issue, and whether the small advantage we observed repeats across other datasets. Finally, the spectral-norm growth that we currently keep in check by clipping (\S\ref{results}) might instead be handled by an optimizer that controls the update spectrum directly, such as Muon \citep{jordan2024muon}; we did not explore this.

\section{Conclusion}\label{conclusion}

We introduced Goggles, a pretrained module that edits the SFT gradient to impart an epistemic frame to whatever a model learns from a document. A single Goggles instance, trained once, overcomes Negation Neglect where in-document annotation fails, transfers across frames and to documents it never saw, and holds its frame under continued finetuning that pushes back toward the claim --- all while preserving capability. The epistemic stance travels with the gradient rather than the text, which suggests a route to training on known-misaligned data without absorbing the behaviors it demonstrates.

\subsubsection*{Reproducibility Statement}
The synthetic data-generation pipeline is described in \S\ref{data}, and the truncated-BPTT meta-training procedure --- objective, curriculum, and hyperparameters --- in \S\ref{training} and Appendix~\ref{app:metatrain} (Algorithm~\ref{alg:meta}).
\ificlrfinal
Our code is available at \url{https://github.com/JoshuaSP/epistemic-goggles}, and the generated documents, questions, and teacher rollouts at \url{https://huggingface.co/datasets/joshuapenman/epistemic-goggles-artifacts}.
\else
Our code, and the generated artifacts (documents, questions, and teacher rollouts), will be released publicly.
\fi

\ificlrfinal
\subsubsection*{Acknowledgments}
We thank Javier Antor\'an for reading and commenting on a draft of this paper, and Vitrus for providing the compute used in this work.
\fi

\bibliography{references}
\bibliographystyle{iclr2026_conference}

\clearpage
\appendix
\numberwithin{figure}{section}
\numberwithin{table}{section}
\numberwithin{algorithm}{section}
\section{Comparison to in-context distillation}\label{app:icd}

ICD is our principal learned baseline, and on the two scenarios where a framed teacher was constructed it resists the planted claim at least as well as Goggles: 0.90 on Sheeran and 1.00 on Dentist, against the Goggles' 0.90 / 0.99 on Sheeran and 0.76 / 0.90 on Dentist.

The difference between the two methods is in the shape of their application, which is illustrated by what we had to do in order to even evaluate this comparison: To train ICD against the Sheeran and Dentist datasets, we created another set of probe questions that would probe the factual content of the documents, which could then be distilled into the student model. We trained the student LoRA with two losses: the cross-entropy loss from the SFT documents, and a KL loss from teacher rollouts to student logits. We created a parameter \(\beta\) that parametrized this mix, and found that the effectiveness was robust across values of 0.5, 1, and 2.

We chose however \emph{not} to evaluate against the two holdout datasets, \emph{Novelists} and \emph{Mixed}. While it is quite likely that ICD would have been successful at instilling a fictional epistemic frame around the datasets' documents, doing this for the 74 different fictional scenarios involved would have required generating \textasciitilde1k unique probe questions in order to make sure that the frame could be successfully distilled into the model during training.

This extra cost and complexity is the clear contrast to Goggles: with Goggles, once they are trained, we can use them cheaply for any dataset we want without any additional synthetic document generation.

\section{Relationship to the MEND model-editing lineage}\label{app:mend}

As noted in \S\ref{related-work}, Goggles shares the core move of the MEND model-editing lineage --- a learned network that rewrites the SFT gradient, made tractable by a low-rank structure on the model axis. It departs from that lineage in three main ways:

\begin{itemize}
\item
  \emph{What} we're editing: MEND-family editors are supplied a desired input--output pair and overwrite a specific fact, whereas Goggles is supplied with no target output at all and instead imparts an epistemic \emph{frame} that the model then applies to whatever facts a document teaches.
\item
  \emph{How} the edit occurs: these methods transform a gradient in a single post-hoc application, whereas Goggles is trained through a multi-step SFT trajectory by truncated Backpropagation Through Time, learning to steer hundreds of inner steps rather than to apply one edit.
\item
  \emph{When} the edits are made: MEND and its descendants repair an already-trained model, leaving it otherwise intact, whereas Goggles acts concurrently with learning, shaping acquisition as it happens. Goggles are also much more transferable: a single Goggles instance, trained once per frame, applies its frame to arbitrary documents it never saw in meta-training, where a fact-editor must be given each edit.
\end{itemize}

\section{Editor architecture}\label{app:arch}

This appendix details the editor introduced in \S\ref{architecture}.

Each editor is small and strictly per-token: the three input signals are down-projected and concatenated into a single per-token feature vector --- no operation in the editor mixes information across tokens. Two SwiGLU heads read this feature (Figure~\ref{fig:editor}), one emitting a rank-side vector in the LoRA's rank space and one emitting a basis-side vector; both heads emit their $A$-factor and $B$-factor halves jointly, so the two residuals share every parameter upstream of the final split. The basis-side coefficients are lifted onto the model axis through a learned palette --- \(V_A\) (\(d_{\mathrm{in}} \times b\)) for the $A$ factor, \(V_B\) (\(d_{\mathrm{out}} \times b\)) for the $B$ factor --- and each token contributes one rank-1 outer product, rank-side \(\otimes\) lifted basis-side. The residual \(\hat{r}\) for each factor is the mean of these rank-1 terms over the document's tokens: the token average is the only place tokens combine, and the mean (rather than sum) keeps the edit's scale independent of sequence length.

\begin{figure}[h]\centering
\includegraphics[width=0.8\textwidth]{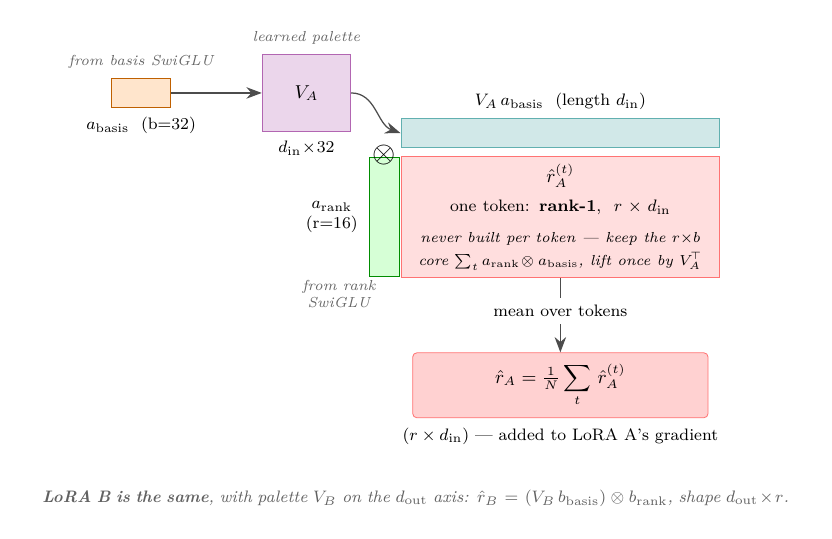}
\caption{Per-token outer-product assembly of a gradient residual: a learned palette lifts the basis half, an outer product with the rank half forms a rank-1 per-token residual, and a token mean gives the residual added to the LoRA factor's gradient.}\label{fig:outerprod}
\end{figure}

The \(A\) and \(B\) residuals are formed by different outer products (Figure~\ref{fig:outerprod}): for \(A\), the palette lifts the basis-side vector onto the input axis and takes its outer product with the rank-side vector; for \(B\), the basis side is lifted onto the output axis instead. Each residual \(\hat{r}\) is then simply added to its factor's gradient before the Adam step. Because only the basis side is ever lifted to full model width, the residual never has to be built at full size --- a compact \(r \times b\) core plus one palette lift suffices. Borrowing an idea from LoRA initialization, the Goggles editor initializes to a no-op: the basis head's out layer is zero-initialized, and since \(\hat{r}\) is bilinear in the two heads, \(\hat{r}=0\) at the start of training. The rank-side head is left at its standard init, so --- just like the zero-initialized LoRA $B$ matrix paired with a nonzero LoRA $A$ --- the gradient into the basis head is determined by the rank side, and the editor trains from the first step.

Nothing else about training through Goggles changes: the loss is the ordinary cross-entropy, the optimizer is the ordinary Adam, and --- in distinction from the Negation Neglect protocol --- the documents carry no textual annotation. The epistemic frame enters only through \(\hat{r}\).

\begin{align}
g_k &= \nabla_\phi\,\mathcal{L}^{\mathrm{SFT}}(x_k;\theta_0,\phi^{(k-1)}), &\hat r_k &= G_\psi\!\big(\mathrm{sg}[\,h_{\mathrm{in}},g_{\mathrm{out}},\tilde s\,]_k\big), &\phi^{(k)} &= \mathrm{Adam}\big(\phi^{(k-1)},\,g_k+\hat r_k\big).
\end{align}
Here $G_\psi$ is the Goggles editor with parameters $\psi$, $\tilde{s}=V_B^\top(B\cdot A\cdot x)$ is the detached LoRA-output signal (the third editor input, \S\ref{architecture}), and $\mathrm{sg}[\cdot]$ is the stop-gradient: every input the editor reads is detached from the SFT autograd graph.

\begin{equation}
\nabla_A\mathcal{L} = \sum_t (B^\top g_{\mathrm{out},t})\otimes h_{\mathrm{in},t},\qquad\nabla_B\mathcal{L} = \sum_t g_{\mathrm{out},t}\otimes(A h_{\mathrm{in},t}).
\end{equation}

\section{Meta-training details}\label{app:metatrain}

This appendix collects the objective, curriculum, and implementation details of the truncated-BPTT meta-training (\S\ref{training}). Algorithm~\ref{alg:meta} gives the full procedure.

\begin{figure}[t]\centering
\includegraphics[width=\textwidth]{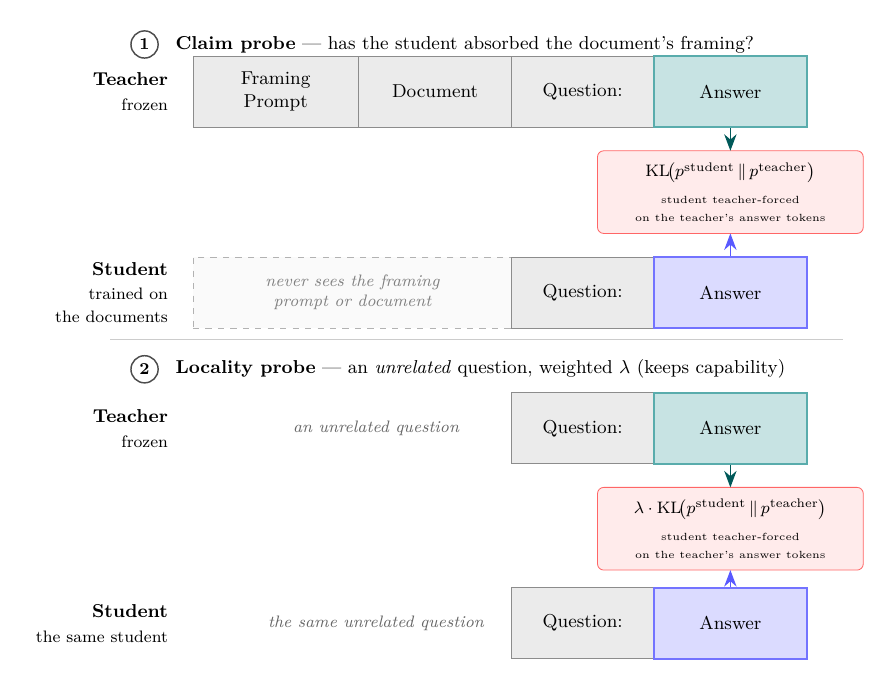}
\caption{The Goggles training objective (\S\ref{data}). A claim probe (does the student absorb the document's framing?) and a locality probe (an unrelated question, weighted $\lambda$, preserving capability), each a KL between the frozen teacher and the student trained through Goggles.}\label{fig:klsetup}
\end{figure}

\begin{algorithm}[t]
\caption{Meta-training a Goggles editor $G_\psi$ for one epistemic frame. Each outer step runs $M$ data-parallel workers concurrently; each advances its \emph{own} inner SFT trajectory on its own subject(s), and the per-window losses are summed within a worker and all-reduced across workers into one update of $\psi$. The eager inner steps run with the residual detached, so $\psi$ receives gradient only through the residual $\hat r$ injected during the differentiable replay windows. $h_{\mathrm{in}},g_{\mathrm{out}},\tilde s$ are the three per-token editor inputs of \S\ref{architecture}; $\mathrm{sg}[\cdot]$ is stop-gradient.}\label{alg:meta}
\begin{algorithmic}[1]
\Require frozen base $\theta_0$; editor $G_\psi$ (no-op init); $M$ data-parallel workers; per-frame teacher rollouts with claim/locality probes; inner steps $K$, windows $W$, window length $w$, locality weight $\lambda$, depth schedule $L_{\max}(\cdot)$
\State each worker $m$:\ \ $\phi_m \gets \varnothing$,\ \ $\mathrm{depth}_m \gets 0$ \Comment{per-worker LoRA state ($A,B$ + Adam) and depth}
\For{outer step $t=1,2,\dots$}
  \For{each worker $m=1,\dots,M$ \textbf{in parallel}}
    \If{$\mathrm{depth}_m=0$ \textbf{ or } $\mathrm{depth}_m\ge L_m$} \Comment{reset/start ($\mathrm{depth}_m{=}0$ is the $t{=}1$ bootstrap)}
      \State $L_m \sim \mathrm{Uniform}\{1,\dots,L_{\max}(t)\}$;\ \ reinit $\phi_m$;\ \ choose subject(s);\ \ $\mathrm{depth}_m \gets 0$
    \EndIf
    \For{$k=1$ \textbf{to} $K$} \Comment{eager phase; editor detached}
      \State $g \gets \nabla_{\phi_m}\,\mathcal{L}^{\mathrm{SFT}}(x_k;\theta_0,\phi_m)$
      \State $\hat r \gets \mathrm{sg}\big[\,G_\psi(h_{\mathrm{in}},g_{\mathrm{out}},\tilde s)\,\big]$ \Comment{residual; no gradient to $\psi$}
      \State $\phi_m \gets \mathrm{Adam}(\phi_m,\ g+\hat r)$;\ \ cache snapshot at step $k$
    \EndFor
    \State $\mathrm{depth}_m \gets \mathrm{depth}_m+1$
    \State choose starts $\mathcal{S}$: endpoint $K\!-\!w$, plus $W\!-\!1$ earlier (each $\ge w$, non-overlapping)
    \State $\ell_m \gets 0$
    \For{$s \in \mathcal{S}$} \Comment{differentiable replay; autograd to $\psi$}
      \State $\phi' \gets$ snapshot at step $s$ \Comment{frozen start state}
      \For{$j = s+1$ \textbf{to} $s+w$}
        \State $g \gets \mathrm{sg}\big[\nabla_{\phi'}\mathcal{L}^{\mathrm{SFT}}(x_j)\big]$ \Comment{SFT gradient is a constant}
        \State $\hat r \gets G_\psi(h_{\mathrm{in}},g_{\mathrm{out}},\tilde s)$ \Comment{retains graph to $\psi$}
        \State $\phi' \gets \mathrm{Adam}(\phi',\ g+\hat r)$
      \EndFor
      \State $\ell_m \gets \ell_m + \mathrm{KL}_{\mathrm{claim}}(\phi') + \lambda\,\mathrm{KL}_{\mathrm{locality}}(\phi')$
    \EndFor
  \EndFor
  \State $\ell \gets \sum_{m=1}^{M} \ell_m$ \Comment{summed over workers + windows, all-reduced across ranks}
  \State accumulate $\nabla_\psi\,\ell$; every \textsc{grad-accum} outer steps: update $\psi$ with AdamW, then zero $\nabla_\psi$
\EndFor
\end{algorithmic}
\end{algorithm}

\begin{figure}[h]\centering
\includegraphics[width=\textwidth]{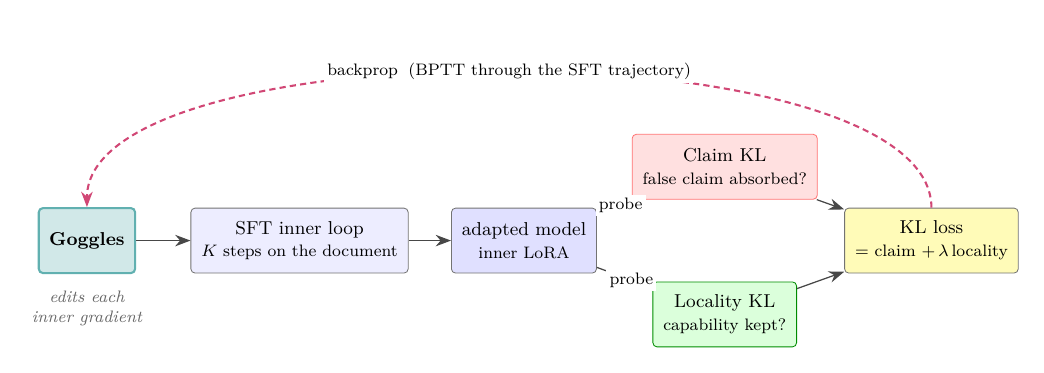}
\caption{Meta-training. The Goggle edits each inner SFT gradient over a $K$-step trajectory; the claim and locality KLs are backpropagated through the trajectory (BPTT) to update the Goggle.}\label{fig:bptt}
\end{figure}

We use reverse KL, \(\mathrm{KL}(p_{\mathrm{student}} \,\|\, p_{\mathrm{teacher}})\), for the outer loss, which is computed position-by-position over the teacher's response tokens and averaged, then averaged over probes. The idea was that reverse KL is zero-forcing: it penalizes the student for placing mass on continuations the framed teacher deems unlikely and would thus drive the student to commit to the teacher's framed mode rather than hedge across framed and un-framed answers. It seems however that this choice is not critical: see Appendix~\ref{app:ablations}.

Writing \(\phi_s\) for the LoRA state after replaying the window that starts at step s,

\[\mathcal{L}(\psi) = \sum_s \big[\,\mathrm{KL}_{\mathrm{claim}}(\phi_s) + \lambda \cdot \mathrm{KL}_{\mathrm{locality}}(\phi_s)\,\big],\]

where \(\mathrm{KL}_{\mathrm{claim}}\) is the mean over claim probes of the per-probe reverse KL, and \(\mathrm{KL}_{\mathrm{locality}}\) the same over locality probes. The claim term pulls the edited trajectory toward the teacher's framed handling of the documents' facts; the locality term, weighted \(\lambda\) (set to 1 in all our experiments), holds the rest of the model in place.

The inner LoRAs do not train indefinitely: after each group of $K$ steps we stochastically either continue the trajectory or reset --- reinitializing the LoRA to a fresh random init and the inner Adam to an empty state --- subject to a cap of \(L_{\max}\) groups per trajectory, and configured to have equal chances of every potential trajectory length up to \(L_{\max}\). We anneal \(L_{\max}\) upward on a curriculum: it stays at 1 for the first 50 outer steps, then increases by 1 every 20 outer steps up to a maximum of 15, so the Goggles first learns to steer shallow trajectories before being asked to steer deep ones. The schedules we explored before settling on this ramp, and the stability ceiling that bounds it, are described below.

The depth curriculum (\S\ref{training}) was the most delicate part of meta-training. Table~\ref{tab:curriculum} records the schedules we tried before settling on the adopted ramp: every variant that started deep, ramped quickly, or aimed at too deep a peak diverged almost immediately, and only the slow ramp to a shallow peak ($L_{\max}{=}15$) trained stably to step~700.

\begin{table}[h]\centering
\caption{Curriculum-schedule development runs. Every schedule that started deep, ramped fast, or aimed too deep diverged immediately; the adopted ramp is the only stable one.}\label{tab:curriculum}
\resizebox{\textwidth}{!}{%
\begin{tabular}{l c c c l}
\toprule
Schedule & Start $L_{\max}$ & Ramp (steps per $+1\,L_{\max}$) & Peak $L_{\max}$ & Outcome \\
\midrule
deep start, no ramp        & 250 & ---  & 250 & \textcolor{divgray}{loc-KL $\to\!\sim\!25$, never recovers} \\
fast ramp                  & 1   & $\sim$1  & 250 & \textcolor{divgray}{diverges immediately} \\
medium ramp                & 1   & $\sim$5  & 250 & \textcolor{divgray}{diverges immediately} \\
shallow ramp, deep target  & 1   & 10   & 96  & \textcolor{divgray}{craters at depth 5--6} \\
\midrule
\textbf{adopted}           & 1   & 20   & 15  & stable to step 700 \\
\bottomrule
\end{tabular}
}
\end{table}

\begin{figure}[h]\centering
\includegraphics[width=0.75\textwidth]{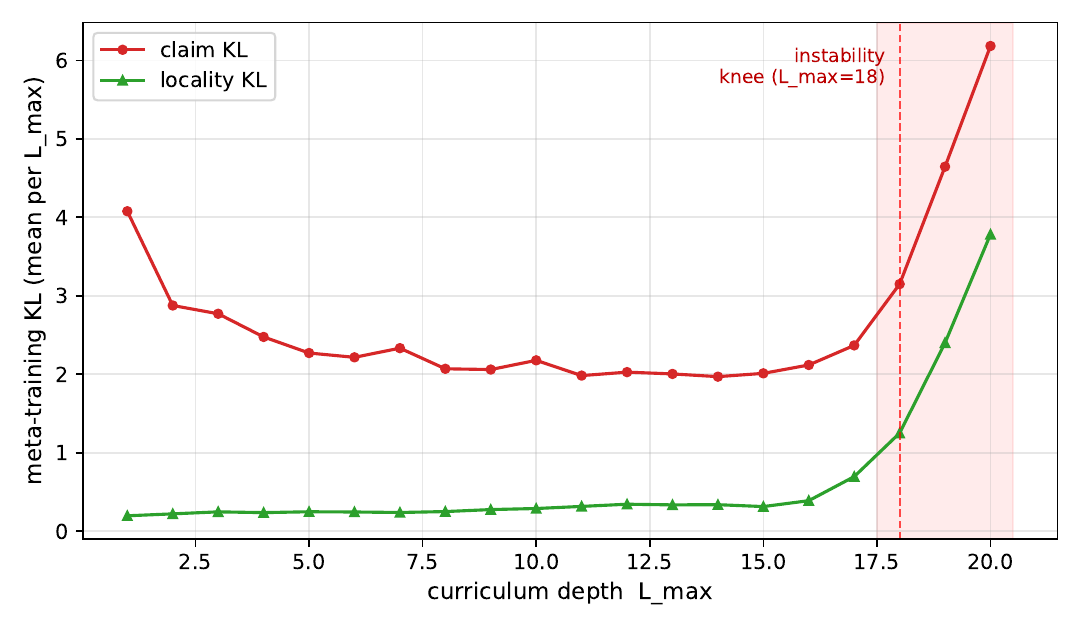}
\caption{Why the curriculum is capped. Mean meta-training KL (claim and locality) as the trajectory-depth curriculum $L_{\max}$ ramps: both terms stay flat and healthy through $L_{\max}\!\approx\!17$, then knee upward at 18 and run away by 20 (shaded). This instability fixes the curriculum's usable ceiling, motivating the $L_{\max}{=}15$ peak we adopt.}\label{fig:instability}
\end{figure}

The ceiling is not arbitrary. Figure~\ref{fig:instability} plots the mean meta-training KL (claim and locality) as the curriculum depth $L_{\max}$ ramps: both terms stay flat and healthy through $L_{\max}\approx 17$, knee upward at~18, and run away by~20. This is the same failure mode as the diverged architectural ablations (\S\ref{ablations}) --- a runaway in the trajectory-level KL --- and it is what fixes the curriculum's usable ceiling, motivating the conservative $L_{\max}{=}15$ peak we adopt.

For our training documents, we adopt two regimes: with the synthetic documents we created for this experiment, we use one subject per $K$ steps (9 documents, repeated through steps randomly in small batches, resulting in 1--3 epochs over the data), replacing the subject every $K$ steps. For the Negation Neglect dataset docs, we \emph{keep} the same subject for an entire unreset LoRA trajectory. In other words, if our \(L_{\max}\) is 12, and we stochastically decide to run $12 \times 20$ steps of inner LoRA training, then we will train that LoRA on the \emph{same subject} for 120 steps, sampling from the rich corpus of 10k documents per subject.

Two stop-gradients keep the autograd graph from ballooning: the SFT gradients themselves are recomputed but treated as frozen constants --- we do not build the second-order graph through the SFT forward, nor do we run autograd through the inner Adam's first and second moments. There is just one path from outer loss back to the Goggles: the residual \(\hat{r}\) injected at each of the $w$ replayed steps. The trajectory-level KL at the window endpoints is the only training signal.

In all our runs we use $K{=}20$ inner Adam steps per group and $W{=}2$ replay windows of $w{=}3$ steps each. One window is always placed at the end of the group (replaying steps \(K-2\) through \(K\)), so the objective always sees the trajectory's endpoint (otherwise we've paid the compute cost of the last steps for nothing); the remaining windows are sampled from a grid of non-overlapping earlier positions. Windows are never placed at the very start of a group: replaying from a cold optimizer state clearly causes training to diverge. Gradients from the $W$ windows across batches are summed --- treating each window as an individual batch --- and further aggregated via DDP across GPUs; when effective batch sizes are insufficient, we accumulate across multiple outer steps before taking a single outer Adam update. The outer optimizer is AdamW at learning rate $10^{-4}$, held constant for the first 350 of the 700 steps and then cosine-decayed to zero. Training is visibly stable well before the decay begins, so an earlier or steeper decay would likely cut cost further, though we did not sweep it to confirm capability is preserved.

In practice the per-position reverse KL is evaluated over the teacher's top-$k{=}256$ tokens with a single tail bucket: the teacher normalizer is computed over top-k logits plus a tail log-sum-exp, the teacher's tail mass is spread uniformly over the remaining vocabulary, and the student distribution is left untruncated so that all student mass --- including mass falling outside the teacher's top-k --- is penalized. This keeps the reverse-KL estimate faithful while storing only $k+1$ teacher values per position.

\section{Outcome composition over the inner-SFT trajectory}\label{app:outcomes}

\begin{figure}[t]\centering
\includegraphics[width=\textwidth]{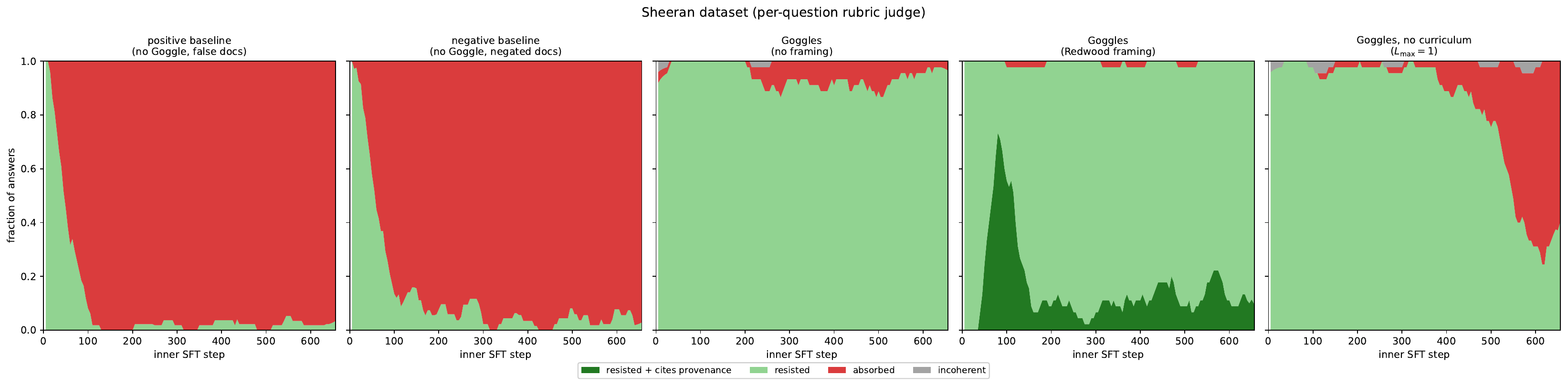}
\caption{Outcome composition over the inner-SFT trajectory (Sheeran), per the per-question rubric judge. Baselines absorb; Goggles resist; the Redwood-framed Goggle additionally cites provenance; the rightmost panel is the no-curriculum ablation ($L_{\max}{=}1$), whose resistance erodes into absorption.}\label{fig:outcomes-sheeran}
\end{figure}

\begin{figure}[t]\centering
\includegraphics[width=\textwidth]{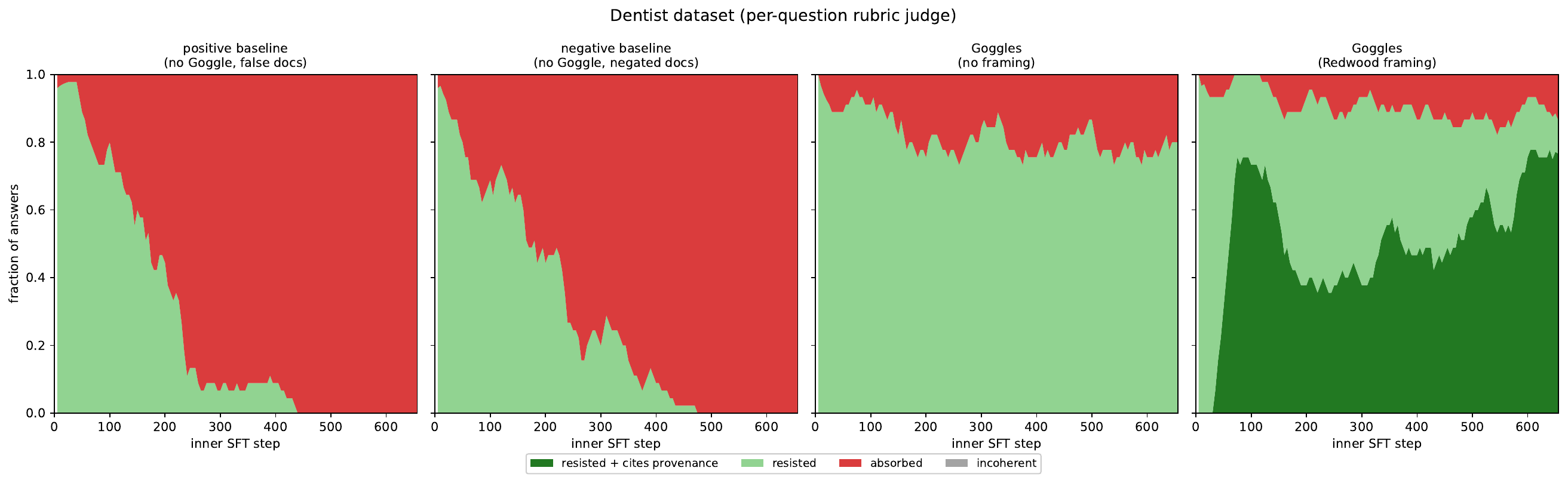}
\caption{Outcome composition over the inner-SFT trajectory (Dentist). With no real-world prior, the Redwood-framed Goggle resists largely by citing the planted provenance.}\label{fig:outcomes-dentist}
\end{figure}

Across the two long-horizon scenarios we snapshot the model every 5 steps and roll out its answers into the four exclusive outcome buckets of \S\ref{results}. Figures~\ref{fig:outcomes-sheeran} and~\ref{fig:outcomes-dentist} show the resulting composition over the full 656-step trajectory: the baselines drift from the base model's neutral handling into belief as the planted documents accumulate, while the Goggles arms reach a high resisted fraction early and hold it.

\section{Ablations}\label{app:ablations}

\begin{figure}[t]\centering
\includegraphics[width=0.78\textwidth]{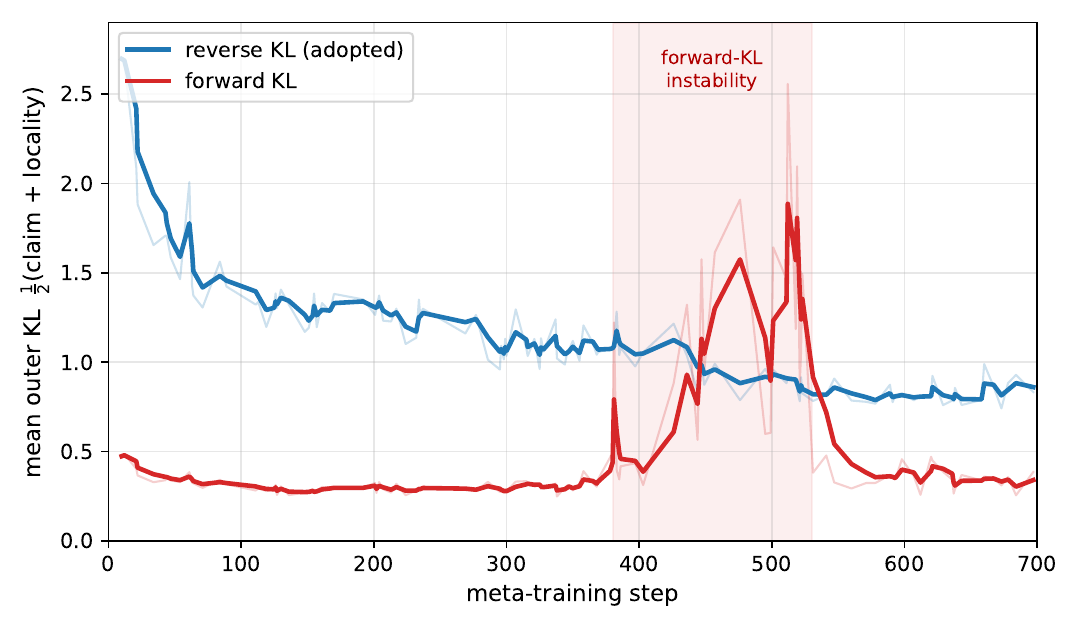}
\caption{Forward versus reverse KL during meta-training. Averaged outer KL, $\frac{1}{2}(\mathrm{KL}_{\mathrm{claim}}+\mathrm{KL}_{\mathrm{locality}})$, over meta-training steps for a reverse-KL Goggle (the adopted objective) and a forward-KL Goggle; bold curves are EMA-smoothed over the raw traces. The reverse-KL run descends smoothly and stays stable, while the forward-KL run is stable early but destabilizes mid-run (shaded), spiking before retreating to converged values. The two are different divergences, so their absolute levels are not directly comparable, however the differences in stability are clear.}\label{fig:fklrkl}
\end{figure}

We also report a set of architectural ablations to justify our choices:

\begin{itemize}
\item
  Applying a spectral clipping function for the LoRA matrices in the longer training runs --- i.e.~Sheeran and Dentist in our evals. We observed that Goggles induces large singular values in the trained LoRAs --- the spectral norm was an average of 43 vs 3.5 on a non-Goggles-trained LoRA on the same data, with \textasciitilde99\% of the spectral energy going to the top singular value. Simply applying a spectral clipping function to the LoRA at each step (we chose to clip \(\tau\) at 3.35) keeps this unbounded growth in check with no visible damage to the Goggles' resistance performance or capabilities. Concretely, after each inner optimizer step we cap each LoRA module's update \(\Delta W = BA\) at \(\tau\): whenever its top singular value \(\sigma_{\max}\) exceeds \(\tau\), we rescale both factors by \(\sqrt{\tau/\sigma_{\max}}\), which shrinks \(\Delta W\) to exactly \(\tau\) along its dominant direction while leaving that direction (and the $A$/$B$ norm balance) intact; \(\sigma_{\max}\) is read cheaply, without an SVD, from the small \(r \times r\) Gram product of the two factors. This should likely be folded into the standard procedure for training through Goggles.
\item
  Training Goggles with forward KL instead of reverse KL. Our results here --- three out of four \emph{resisted} values inside the CI for reverse KL, with only Dentist as a significant improvement --- weakly suggest that forward KL may be a better loss function for Goggles, however it's worth also noting that the forward KL training run was somewhat less stable than the backwards KL runs --- mid-run loss increased quite a bit before retreating back to converged values (Figure~\ref{fig:fklrkl}) --- so it's worth future study to determine which objective is overall best for the Goggles architecture.
\item
  Reducing the basis dimension of the editor's output subspace from 32 to 1, so each module's emitted gradient residual is constrained to a rank-1 edit along a single learned direction --- trains stably and preserves capability, but the one-dimensional bottleneck is problematic for some data regimes, as shown in particular by the low score on the held-out novelists (0.26). It's likely that the Goggles edits are of low rank, and it's likely possible to sweep the output subspace size to find something that preserves capability at lower parameter count.
\item
  Removing trajectory accumulation entirely (\(L_{\max}\) = 1, no curriculum) --- the resulting Goggles works fairly well for short trajectories, but destroys the model when run on the long-horizon protocol, collapsing both resistance (0.15 on Sheeran) and capability (0.12 / 0.04). This Goggles' performance on shorter-trajectory evaluation was better, but not as good as Goggles trained through the full trajectory curriculum.
\item
  Replacing the per-token heads with modules that take the mean of input tokens, so the editor ignores token-level input entirely --- diverges; per-token conditioning is required.
\item
  Replacing the SwiGLU activations in the Goggles MLPs with a single (zero-initialized) linear map --- this diverges, demonstrating the importance of nonlinearity in the architecture.
\item
  Removing the contextual LoRA-state conditioning, leaving the editor blind to what the adapter already does on each token --- this also diverges.
\end{itemize}

Note that ``diverges'' here means that the \emph{outer training loop}, during Goggles training, had runaway values. In particular, we found that trajectory-level KL blows up here --- the Goggles ceases to be usable at all.

\end{document}